%% file: main.tex
\documentclass{article}

\usepackage{microtype}
\usepackage{graphicx}
\usepackage{algorithm}
\usepackage[super]{nth}
\usepackage{booktabs}
\usepackage{subcaption}
\usepackage{textcomp}
\usepackage{pythonhighlight}

\usepackage[colorinlistoftodos,prependcaption,textsize=tiny]{todonotes}

\usepackage{hyperref}

\usepackage[accepted]{icml2020}
\usepackage[titletoc,title]{appendix}
\icmltitlerunning{Are Hyperbolic representations in graphs created equal?}
\input{math_commands}

\begin{document}

\twocolumn[
\icmltitle{Are Hyperbolic Representations in Graphs Created Equal?}




\begin{icmlauthorlist}
\icmlauthor{Max Kochurov}{sk}
\icmlauthor{Sergey Ivanov}{cr}
\icmlauthor{Eugeny Burnaev}{sk}
\end{icmlauthorlist}

\icmlaffiliation{sk}{Skolkovo Institute of Science and Technology, Russia}
\icmlaffiliation{cr}{Criteo AI Lab, France}
\icmlcorrespondingauthor{Max Kochurov}{maxim.v.kochurov@gmail.com}

\icmlkeywords{Machine Learning, ICML}

\vskip 0.3in
]



\printAffiliationsAndNotice{}  

\begin{abstract}
Recently there was an increasing interest in applications of graph neural networks in non-Euclidean geometry; however, are non-Euclidean representations always useful for graph learning tasks? For different problems such as node classification and link prediction we compute hyperbolic embeddings and conclude that for tasks that require global prediction consistency it might be useful to use non-Euclidean embeddings, while for other tasks Euclidean models are superior. To do so we first fix an issue of the existing models associated with the optimization process at zero curvature. Current hyperbolic models deal with gradients at the origin in ad-hoc manner, which is inefficient and can lead to numerical instabilities.  We solve the instabilities of $\kappa$-Stereographic model at zero curvature cases and evaluate the approach of embedding graphs into the manifold in several graph representation learning tasks. 
\end{abstract}

\section{Introduction}
Hierarchies in data are common in real world settings and can be observed in many scenarios. For example, languages have relations between words and contextual dependencies within a sentence. Words can be viewed as entities and one may define natural type dependencies on them. These dependencies may be entirely arbitrary and different relation graphs may exist for the same dictionary.

In general, there are several settings in graph problems: knowledge graph completion~\cite{mrelpoincare2019balazevic}, node classification and link prediction~\cite{hgnn2019liu,hgcn2019chami,constcur2019bachmann} or graph embedding~\cite{mixedcur2019paszke}. While Euclidean baselines are strong, there is a general trend of making Hyperbolic embeddings to be more efficient and interpretable.

In this work we highlight the shortcomings of the past research about Hyperbolic deep learning for graph data. We review the model from \cite{constcur2019bachmann} and fix it to be more robust at zero curvature case. Furthermore, we evaluate the $\kappa$-Stereographic model for node classification, link prediction, graph classification, graph embedding problems.

\section{Preliminaries}
\subsection{Base Model}
\begin{figure}[b]
    \begin{subfigure}{0.5\linewidth}
    \centering
    \includegraphics[width=0.7\linewidth]{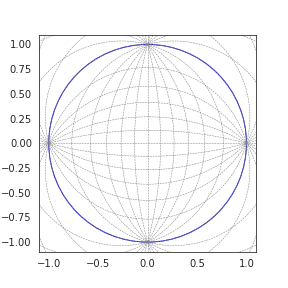}
    \caption{$\kappa=1$}
    \label{fig:geodesics-sproj-1}
    \end{subfigure}%
    \begin{subfigure}{0.5\linewidth}
    \centering
    \includegraphics[width=0.7\linewidth]{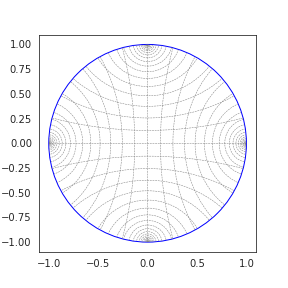}
    \caption{$\kappa=-1$}
    \label{fig:geodesics-sproj-2}
    \end{subfigure}
    \caption{Stereographic projection geodesics for sphere ($\kappa=1$) and hyperboloid ($\kappa=-1$)}
    \label{fig:geodesics-sproj}
\end{figure}
$\kappa$-Stereographic model $\gM_\kappa^n$~\citep{constcur2019bachmann} is a unification of constant curvature manifolds: hyperboloid and sphere. The model is a Riemannian manifold that has constant sectional curvature $\kappa$ and dimension $n$. Besides curvature, the parameter $\kappa$ defines how parameters are constrained what is crucial for optimization algorithms:

\begin{equation}
    \gM_\kappa^n = \begin{cases}
    \left\{x \in \sR^n : \|x\|_2 < 1/\sqrt{-\kappa}\right\},\; \lambda^\kappa_x \qquad &\kappa < 0\\
    \sR^n,\; \lambda^\kappa_x \qquad &\kappa \ge 0
    \end{cases}\label{eq:manifold}
\end{equation}
where $\lambda^\kappa_x = \frac{2}{1 + \kappa \|x\|_2^2}$ is conformal (preserving angles) metric tensor at point $x$.

For optimization we need exponential map (gradient update) and parallel transport (momentum update). For positive and negative cases they are well defined. Exponential map is a function that defines a unit time travel along geodesic (straight) line from a given point, i.e. $\exp^\kappa_x : T_x\gM_\kappa^n \mapsto \gM_\kappa^n$. For $\kappa$-Stereographic model exponential map is defined as follows:
\begin{equation}
    \exp^\kappa_x(u) = x\oplus_\kappa \tan_\kappa(\|u\|_x/2) \frac{u}{\|u\|_2},
\end{equation}
where $\oplus_\kappa$ is defined as
\begin{equation}
    x \oplus_\kappa y =
        \frac{
            (1 - 2 \kappa \langle x, y\rangle - \kappa \|y\|^2_2) x +
            (1 + \kappa \|x\|_2^2) y
        }{
            1 - 2 \kappa \langle x, y\rangle + \kappa^2 \|x\|^2_2 \|y\|^2_2
        }
\end{equation}
and $\tan_\kappa$ as 
\begin{equation}
    \tan_\kappa(x) = \begin{cases}
    \kappa^{-1/2}\tan (x \kappa^{1/2})& \kappa > 0\\
    x & \kappa = 0\\
    |\kappa|^{-1/2}\tanh (x |\kappa|^{1/2})& \kappa < 0\\
    \end{cases}
    \label{eq:tank}
\end{equation}
Parallel transport (e.g. for momentum update in SGD) along geodesic looks like
\begin{equation}
    P^\kappa_{x\to y}(v)
        =
        \operatorname{gyr}[y, -x] v \lambda^\kappa_x / \lambda^\kappa_y,
\end{equation}
where $\operatorname{gyr}[u, v]w$ is defined as 
\begin{equation}
    \operatorname{gyr}[u, v]w
        =
         (- (u \oplus_\kappa v)) \oplus (u \oplus_\kappa (v \oplus_\kappa w))
\end{equation}
Computing distances and similarity measured on the $\kappa$-Stereographic manifold is used for final classification layers in neural networks. Importantly, distances on the manifold are defined as follows:
\begin{equation}
    d_\kappa(x, y) = 2\tan_\kappa^{-1}(\|(-x)\oplus_\kappa y\|_2),
\end{equation}
Gromov product, an extension to inner product on manifolds
\begin{equation}
    (x, y)_r = (d_{\gM}(x, r)^2 + d_{\gM}(y, r)^2 - d_{\gM}(x, y)^2) / 2,
    \label{eq:gromov-product}
\end{equation}
and distance to the hyperplane
\begin{align}
        d_\kappa(x, \tilde{H}_{a, p}^\kappa)
        &=
        \inf_{w\in \tilde{H}_{a, p}^\kappa} d_\kappa(x, w)\\
        &=
        \sin^{-1}_\kappa\left\{
            \frac{
            2 |\langle(-p)\oplus_\kappa x, a\rangle|
            }{
            (1+\kappa\|(-p)\oplus_\kappa \|x\|^2_2)\|a\|_2
            }
        \right\},
        \label{eq:distance-to-hyperplane}
    \end{align}
where 
\begin{equation}
    \sin_\kappa(x) = \begin{cases}
    \kappa^{-1/2}\sin (x \kappa^{1/2})& \kappa > 0\\
    x & \kappa = 0\\
    |\kappa|^{-1/2}\sinh (x |\kappa|^{1/2})& \kappa < 0\\
    \end{cases}
    \label{eq:sink}
\end{equation}
\subsection{Fixing Missing Gradients}
Equations~\ref{eq:tank} and \ref{eq:sink} are incomplete. At the point of zero curvature, gradient for $\kappa$ is not well defined. However, we can generalize it by taking left and right Taylor expansions. We can see gradients for $\kappa$ appear:
\begin{align}
    \tan_\kappa(x) &\approx
    x
    + \tfrac{1}{3} \kappa  x^3%
    + \tfrac{2}{15} \kappa^2 x^5%
    \Big|_{\kappa=+\varepsilon}%
    \\
    \tan_\kappa(x) &\approx
    x
    + \tfrac{1}{3} (-\kappa)  x^3%
    + \tfrac{2}{15} (-\kappa)^2 x^5%
    \Big|_{\kappa=-\varepsilon}
\end{align}
Then the complete formula in the equation \ref{eq:tank} for $\tan_\kappa(x)$ that is differentiable at $\kappa=0$ should be the following:
\begin{equation}
    \tan_\kappa(x) = \begin{cases}
    \kappa^{-1/2}\tan (x \kappa^{1/2})& \kappa > 0\\
    x
    + \tfrac{1}{3} \kappa  x^3%
    + \tfrac{2}{15} \kappa^2 x^5%
    & \kappa = 0\\
    |\kappa|^{-1/2}\tanh (x |\kappa|^{1/2})& \kappa < 0\\
    \end{cases},
\end{equation}
where $\tfrac{2}{15} \kappa^2 x^5$ term is optional and is required for higher order gradients only. Taylor expansions for $\tan^{-1}_\kappa$ and functions involved in computing distances to hyperplanes may be found in Appendix~\ref{app:taylor}.

\subsection{Curvature optimization\label{sec:curvature-optimization}}
Curvature optimization in Hyperbolic models is an important challenge. The optimization step changes the geometry of space and parameter constraints are also dependent on the curvature parameter. Therefore parameters and curvature are tied together and curvature update cannot be performed without parameter updates.

We are interested in optimizing parameters $\{p_i\}$ of the model that lie in $\kappa$-Stereographic model. Formally, the optimization problem looks as:
\begin{equation}
    \min_{\kappa\in\sR,\;p_i\in \gM_\kappa^n} \gL(\kappa, \{p_i\})
\end{equation}

To work with parameters $\{p_i\}$ on the manifold, they are represented as tuples of numbers in some chart, i.e. a numerical parametrization of the manifold. One cannot make independent parameter updates. Fair to note, that all curvatures represent the same model, and one can be obtained with isomorphism. However, this is not the case for product manifolds and numerical precision is different for different curvatures \cite{mixedcur2019paszke, representation2018desa}.

Model decomposition changes loss landscape and this was shown to improve the performance~\cite{weightnorm2016salimans}. Parameter constraints vary with $\kappa$ as seen in equation~\ref{eq:manifold}.
Once negative $\kappa$ changes, numerical representation of parameters may no longer satisfy boundary conditions, because the ball's radius changes (\eqref{eq:manifold}), while parameters remain fixed. In order to satisfy the constraints, we should project these parameters back to the domain. There are a few alternatives on how to perform optimization in such a complex parameter space.

\paragraph{Alternating optimization.} One way to perform optimization is by doing alternation in parameter updates: with one step we optimize $\kappa$ and project $p_i$, with another step we optimize $p_i$. This partially solves inconsistency but introduces another problem. After $\kappa$ update, all mutual distances are updated and perturbed. Therefore, parameter distances to the origin are changed and momentum for these parameters correspond to a new tangent space, which does not correspond to the old one. From one point of view, it is just a vector space around a specific point; from another, it is connected to a point on a concrete manifold with the metric tensor. After curvature updates, the metric tensor changes as well. It is not obvious how to take this into account. However, just ignoring both problems worked sufficiently in practice.

\paragraph{Joint optimization.} The purpose of alternating optimization was to decouple curvature and parameters. However, it is not practical: forward model pass is done twice, one per each parameter group. More efficiently we can compute and apply gradients once for all parameter groups. There are two options on how to do that: one is to update $\kappa$ first and then $p_i$, the second is vice-versa. In the former case, we have biased gradients for parameters, in the latter for curvature.
Bias appears as we separate updates in two blocks. In parameter updates, curvature affects the exponential map and gradient norms. For curvature updates, we make parameter updates with one curvature and then change curvature without additional computing gradient at a new point. As there are much fewer parameters involved in the curvature updates, biasing curvature in update is more accurate.

\paragraph{Tangent space optimization.} The most consistent way to formulate optimization problems is to reparametrize parameter space and remove constraints~\cite{triviopt2019lezcano}. The intuitive way to do that is to put the problem into the tangent space of zero. The problem formulation changes as follows:
\begin{align}
   \min_{\kappa\in\sR,\;p_i\in \gM_\kappa^n}  \gL(\kappa, \{p_i\}) \quad\Longleftrightarrow \\ 
    \Longleftrightarrow \min_{\kappa\in\sR,\;\tilde p_i\in T_0\gM_\kappa^n} \gL(\kappa, \{\exp_0^\kappa(\tilde p_i)\})
    \label{eq:reparam-tangent}
\end{align}

Next, we compare joint optimization and tangent space optimization with curvature in an empirical study. We start with embedding nodes of a graph into non-Euclidean space such that it preserves graph distances. This problem isolates optimization process from the architecture choice and serves as a good example to study various optimization techniques in non-Euclidean space.

\section{Experiments}

We consider four benchmark tasks: 1) node classification, 2) link prediction, 3) graph classification, 4) graph embedding. These tasks are conceptually different: in tasks 1) and 3) target label for an object does not depend on other objects such as other nodes or graphs. In contrast, to solve problems 2) and 4) it is important to keep global consistency. For link prediction it prevents spurious connections or their absence. For graph embedding task, the loss function captures all node interactions.
\subsection{Graph Embedding\label{sec:graph-embeddings}}
As shown in~\cite{representation2018desa}, Euclidean space cannot capture all tree-like data structures, while embeddings in Hyperbolic space can represent any tree without significant distortions. Other works~\cite{mixedcur2019paszke} successfully applied mixed curvature spaces for graph embeddings and show that non-Euclidean spaces are good at capturing relations in the real-world graphs. The analysis of mixed curvature spaces is still not explored to a full extent. Curvature signs should be specified before training and cannot be changed while training. The limitation requires a costly grid search to tune hyperparameters.
\input{tables/graph_embedding}
The problem is not self-contained to serve as a real-world example, but it clearly shows the advantages and disadvantages of Riemannian optimization compared to Euclidean optimization in terms of the number of iterations and running time. It allows us to compare purely the methods of optimization and the loss function rather than the complexity of neural networks.

In this work, we propose to use a $\kappa$-Stereographic model with smooth transitions in geometry. Tunable curvature allows only to specify the desired product space for embeddings without the sign. Gradient descent is supposed to fit optimal curvature for the graph dataset. We investigate curvature optimization approaches described in Section~\ref{sec:curvature-optimization} for Facebook social network dataset. To embed a graph into metric space, we choose the following loss function (see also Algorithm~\ref{alg:graph-embedding})
\begin{equation}
    D_{avg} = \E_{v_i, v_j \sim G}\left(\frac{d_{ij}}{d_\gM(v_i, v_j)} - 1\right)^2 \to \min_{v_i, v_j}\label{eq:Davg}
\end{equation}
The loss function optimizes the ratio between distances computed on the graph and in the metric space. When ideally optimized, the loss should be strictly zero. The results for four optimization approaches are presented in Table~\ref{tab:fb-graph}.
\input{tables/link_prediction}
\subsection{Node Classification}


In this problem we project embeddings to $\kappa$-Stereographic manifold with a GCN model\footnote{Reference implementation is in found in \texttt{torch\_geometric} package~\cite{torch_geometric2019fey}: \url{https://github.com/rusty1s/pytorch_geometric/blob/master/examples/gcn.py}} described in~\cite{gnn2017kipf}. The final layer to obtain logits was combined with Gromov product (\eqref{eq:gromov-product}), where the reference point is chosen to be zero. Gromov product is a natural generalization of inner products traditionally used for similarity measure purposes. The algorithm for node classification may be found in Algorithm~\ref{alg:node-gcn}. As can be seen from the Table~\ref{tab:node-classification} complex geometry did not lead to improvement and the Euclidean model performed better on hold out dataset partition. The explanation may be in the locality of information as no hierarchy is needed to express the solution, and additional degrees of freedom leads to overfitting.

\input{tables/node_classification}

\subsection{Link Prediction}

We were inspired by \cite{gnn2017kipf} and extended GCN models to work in $\kappa$-Stereographic space. We embed graph neural network in $\kappa$-Stereographic space using the exponential map and measure similarity with Gromov Product. Experiment setup followed standard protocol as in reference implementation in \texttt{torch\_geometric}\footnote{found in \url{https://github.com/rusty1s/pytorch_geometric/blob/master/examples/link_pred.py}}. Evaluation results are found in Table~\ref{tab:link-prediction}. Results suggest that non-Euclidean geometry of embeddings improves generalization across three benchmark datasets. No additional hyperparameters were introduced except tunable curvature. We believe this result is thanks to metric learning nature of the problem, GCN learned to accurately predict such embeddings for nodes that capture the unobserved topological structure.

\subsection{Graph Classification}


For benchmarking classification problems, we studied the Proteins dataset win a modification of GCN (see Algorithm~\ref{alg:graph-gcn}). We embedded graph into $\kappa$-Stereographic model and used Gromov inner product or distance to hyperplanes to compute logit scores for the classification task. The results are presented in Table~\ref{tab:graph-classification}. The results suggest that for graph classification task using non Euclidean embedding may lead to overfitting and training is thus less stable.

\input{tables/graph_classification}

\section{Discussion}
Hyperbolic neural networks are a promising approach for tasks, where knowledge of the global structure is crucial for the prediction. That covers many graph representation learning problems such as link prediction or graph embedding task. From the experiments, we can conclude that more complicated models not necessarily perform better in the problems, where a local structure is enough for solving a task, Euclidean methods are as good as their extensions to non-Euclidean spaces. 

Putting aside optimization of curvature for graph models, one should decide about the nature of the problem at hand and ask whether the answers to the problem depend on all the nodes, subset of them, or just a single object? Often the ground truth labels depend only on the local neighborhood of the node in which case complex Hyperbolic models may be inferior to their Euclidean counterparts.

\bibliography{main.bib}
\bibliographystyle{icml2020}
\newpage
\onecolumn
\begin{appendices}
\input{appendix.tex}
\end{appendices}
\end{document}

%% file: math_commands.tex
\usepackage{amsmath,amsfonts,bm}

\def\eqref#1{equation~\ref{#1}}

\def\1{\bm{1}}

\DeclareMathAlphabet{\mathsfit}{\encodingdefault}{\sfdefault}{m}{sl}
\SetMathAlphabet{\mathsfit}{bold}{\encodingdefault}{\sfdefault}{bx}{n}

\def\gG{{\mathcal{G}}}

\def\gL{{\mathcal{L}}}
\def\gM{{\mathcal{M}}}

\def\sI{{\mathbb{I}}}

\def\sR{{\mathbb{R}}}

\def\sU{{\mathbb{U}}}

\newcommand{\E}{\mathbb{E}}



%% file: tables/graph_embedding.tex
\begin{table}[b]
    \centering
    \caption{Embedding nodes of a graph into $\kappa$-Stereographic space. Results for Facebook dataset $D_{avg}$ ($\downarrow$) (number of gradient descent iterations in brackets). Traditional Euclidean model (\nth{1} row) is outperformed once $\kappa$-Stereographic embeddings are trained on top (\nth{2} row). Hyperbolic Riemannian optimization suffers from week convergence (\nth{3} row), while tangent optimization (\nth{4} row) is able to improve the results given same number of iterations. Time per iteration for all the models is about the same.}
    \label{tab:fb-graph}
    \begin{tabular}{c|c|c}
         \toprule
         Optimization way & $(\gM_\kappa^2)\times 5$ & $(\gM_\kappa^5)\times 2$ \\
         \midrule
         Euclidean (20k) & $0.0069$ & $0.0069$\\
         Euclidean (20k) + $\gM_\kappa^n$ (20k) & $0.0056$ & $0.0055$\\
         $\kappa$-Stereographic (20k) & $0.0085$ & $0.0226$\\
         Tangent $\kappa$-Stereographic (20k) & $0.0075$ & $\mathbf{0.0025}$\\
         \bottomrule
    \end{tabular}
\end{table}

%% file: tables/link_prediction.tex
\begin{table*}[t]
    \caption{Evaluation results (AUC) for link prediction on Cora, PubMed and CiteSeer datasets. Validation partition was used to select test scores to report and 11 independent experiments were run to report standard deviation. 
    }
    \label{tab:link-prediction}

    \centering
    \begin{tabular}{l|c|c|c}
        \toprule
         Models &  Cora & PubMed & CiteSeer\\
         \midrule
         GCN\cite{gnn2017kipf}-$\gM_\kappa^n$-gromov & 
         $\mathbf{0.977\pm0.004}$ & 
         $\mathbf{0.969\pm0.003}$ & 
         $\mathbf{0.990\pm0.002}$\\
         GCN\cite{gnn2017kipf}-$\sR$-inner & 
         $0.864\pm0.01$ & 
         $0.870\pm0.005$ & 
         $0.882\pm0.009$\\
         \bottomrule
    \end{tabular}
\end{table*}

%% file: tables/node_classification.tex
\begin{table}[h]
    \centering
    \caption{Node classification results for GCN model. Accuracy ($\uparrow$) for the model on a test set for best performing model on validation part is used to measure the performance. The results suggest that embedding attributes in $\kappa$-Stereographic space does not lead to improvement.}
    \label{tab:node-classification}
    \begin{tabular}{l|c|c|c}
        \toprule
        Manifold &  Cora & CiteSeer & PubMed\\
        \midrule 
        Baseline $\sR^{16}$ & \textbf{0.809} & \textbf{0.718} & \textbf{0.783} \\
        $\gM_\kappa^{16}$ & 0.8 & 0.68 & 0.76\\
        $(\gM_\kappa^8)\times 2$ & 0.781 & 0.673 & 0.772\\
        $(\gM_\kappa^4)\times 4$ & 0.777 & 0.666 & 0.765\\
        \bottomrule
    \end{tabular}
\end{table}

%% file: tables/graph_classification.tex
\begin{table}[h]
    \centering
    \caption{Graph classification results on PROTEINS dataset. Accuracy ($\uparrow$) is reported for hold out dataset with std.}
    \label{tab:graph-classification}
    \begin{tabular}{l|c|c%
    }
        \toprule
        Model & $D=64$ & $D=32$ 
        \\
        \midrule
        Eucliean baseline & $\mathbf{0.749\pm0.01}$ & $\mathbf{0.795\pm0.02}$
        \\
        $\gM_\kappa^D$-planes & $0.743\pm0.03$&  $0.737\pm0.03$
        \\
        $\gM_\kappa^D$-gromov & $0.735\pm0.03$&  $0.736\pm0.03$
        \\
        \bottomrule
    \end{tabular}
\end{table}

%% file: appendix.tex
\section{Algorithms\label{app:algorithms}}
\input{snippets/graph_embedding}%
\input{snippets/link_prediction}%
\input{snippets/node_classification}%
\input{snippets/graph_classification}%

\newpage

\section{Taylor Expansions\label{app:taylor}}
Proper gradients for zero curvature cases solve all these limitations at once, and no prior knowledge assumed to fit curvature. Moreover, the Tailor series allows us a convenient approach to pre-train Hyperbolic models with zero curvature and, at some point, turn on curvature optimization. Use-cases may involve future research in generalizing Euclidean convolutional neural networks to Hyperbolic spaces. 

The approach to calculating gradients for $\kappa$ correctly is to write tailor expansion at a problematic point.

\begin{align}
    \tan_\kappa(x) &= \begin{cases}
    \kappa^{-1/2}\tan (x \kappa^{1/2})& \kappa > 0\\
    x
    + \tfrac{1}{3} \kappa  x^3
    + \tfrac{2}{15} \kappa^2 x^5
    + \tfrac{17}{315} \kappa^3  x^7
    + \tfrac{62}{2835}  \kappa^4  x^9
    + \tfrac{1382}{155925}  \kappa^5  x^{11} + \dots
    & \kappa = 0\\
    |\kappa|^{-1/2}\tanh (x |\kappa|^{1/2})& \kappa < 0\\
    \end{cases}
    \label{eq:tankfull}
\\
    \tan_\kappa^{-1}(x) &= \begin{cases}
    \kappa^{-1/2}\tan^{-1} (x \kappa^{1/2})& \kappa > 0\\
    x
    - \frac{\kappa  x^3}{3} 
    + \frac{\kappa^2 x^5}{5} 
    - \frac{\kappa^3  x^7}{7} 
    + \frac{\kappa^4  x^9}{9}  
    - \frac{\kappa^5  x^{11}}{11}   + \dots
    & \kappa = 0\\
    |\kappa|^{-1/2}\tanh^{-1} (x |\kappa|^{1/2})& \kappa < 0\\
    \end{cases}
    \label{eq:tankifull}
\\
    \sin_\kappa(x) &= \begin{cases}
    \kappa^{-1/2}\sin (x \kappa^{1/2})& \kappa > 0\\
    x
    - \frac{\kappa  x ^3}{6} 
    + \frac{\kappa^2 x^5}{120} 
    - \frac{\kappa^3  x^7}{5040} 
    + \frac{\kappa^4  x^9}{362880}  
    - \frac{\kappa^5  x^{11}}{39916800}   + \dots
    & \kappa = 0\\
    |\kappa|^{-1/2}\sinh (x |\kappa|^{1/2})& \kappa < 0\\
    \end{cases}
    \label{eq:sinkfull}
\\
    \sin_\kappa^{-1}(x) &= \begin{cases}
    \kappa^{-1/2}\sin^{-1} (x \kappa^{1/2})& \kappa > 0\\
    x
    - \tfrac{1}{6} \kappa  x^3
    + \tfrac{3}{40} \kappa^2 x^5
    - \tfrac{5}{112} \kappa^3  x^7
    + \tfrac{35}{1152}  \kappa^4  x^9
    - \tfrac{63}{2816}  \kappa^5  x^{11} + \dots
    & \kappa = 0\\
    |\kappa|^{-1/2}\sinh^{-1} (x |\kappa|^{1/2})& \kappa < 0\\
    \end{cases}
    \label{eq:sinikfull}
\end{align}

Tailor expansion\footnote{Obtained with Wolfram Mathematica} allows us to take gradient in regions where $\kappa$ is zero. Only first-order expansion is required for the correct gradient and now will depend on the actual value of $x$. With this extension, there is no more gap between constant positive and negative curvature manifolds. They are interpolated smoothly, and gradients allow to determine the best curvature sign.

%% file: snippets/graph_embedding.tex
\begin{algorithm}[h!]
    \caption{Graph embedding\label{alg:graph-embedding}}
    \begin{algorithmic}
    \REQUIRE $G$, $V$ -- graph
    \STATE Run Floyd–Warshall to get graph distances $\{d_{ij}\}$ between nodes
    \STATE Initialize nodes with random embeddigns $\{v_i\} \in \mathcal{M}$\COMMENT{$\gM$ can be any manifold}
    \STATE $D_{avg} = \E_{v_i, v_j \sim G}\left(\frac{d_{ij}}{d_\gM(v_i, v_j)} - 1\right)^2 \to \min_{v_i, v_j}$
    \end{algorithmic}
    %
\end{algorithm}

%% file: snippets/link_prediction.tex
\begin{algorithm}[h!]
    \caption{GCN-$\sU$ for link prediction\label{alg:link-gcn}}
    \begin{algorithmic}
    \REQUIRE $\gG$ -- input graph
    \STATE $\tilde{\gG} = \text{GCN}(\gG)$\COMMENT{Calculate GCN for graph $\gG$}
    \IF {\text{clip norm}}
        \STATE $\forall\:i: y_i= [\tilde\gG]_i / \max(\|[\tilde\gG]_i\|, 1)$\COMMENT{Clip norm of embeddings to 1}
    \ELSE
        \STATE $\forall\:i: y_i=[\tilde\gG]_i$\COMMENT{No norm clipping}
    \ENDIF
    \STATE $\forall\:i: x_i = \exp_0^\gM([\tilde\gG]_i)$\COMMENT{Map GCN output for $i$'th node to $\gM$, e.g. $(\sU^d_1,\dots,\sU^d_K)$}
    \IF {\text{Gromov}}
        \STATE $z^j_i = (x_i, x_j)_0 $\COMMENT{Calculate Gromov product score for link $i$-$j$ as in \eqref{eq:gromov-product}}
    \ELSE
        \STATE $z^j_i = -d_\gM(x_i, x_j)^2 + r $\COMMENT{Calculate distance score for link $i$-$j$ proportional to distance}
    \ENDIF
    \STATE $p(x_i, x_j) = \sigma(z^j_i)$ \COMMENT{Calculate Sigmoid for decision}
    \STATE $\sI(x_i, x_j) = p(x_i, x_j) > t$ \COMMENT{Final decision is based on threshold $t$}
    \end{algorithmic}
\end{algorithm}

%% file: snippets/node_classification.tex
\begin{algorithm}[h!]
    \caption{$\kappa$-GCN for node classification\label{alg:node-gcn}}
    \begin{algorithmic}
    \REQUIRE $\gG$ -- input graph
    \STATE $\tilde{\gG} = \text{GCN}(\gG)$\COMMENT{Calculate GCN for graph $\gG$}
    \STATE $x_i = \exp_0^\gM((\tilde\gG)_i)$\COMMENT{Map GCN output for $i$'th node to $\gM$, e.g. $(\sU^d_1,\dots,\sU^d_K)$}
    \STATE $z^c_i = (x_i, w_c)_0 + b_c$\COMMENT{Calculate Gromov product per class as in \eqref{eq:gromov-product}}
    \STATE $p(x_i) = \operatorname{softmax}(z^1_i, \dots, z^C_i)$ \COMMENT{Calculate Softmax for predictions}
    \end{algorithmic}
\end{algorithm}

%% file: snippets/graph_classification.tex
\begin{algorithm}[h!]
    \caption{$\kappa$-GCN for graph classification\label{alg:graph-gcn}}
    \begin{algorithmic}
    \REQUIRE $\gG$ -- input graph
    \STATE $\tilde{g} = \text{GCN}(\gG)$\COMMENT{Calculate GCN aggregation for graph $\gG$}
    \STATE $x = \exp_0^\gM(\tilde{g})$\COMMENT{Map GCN output for graph to $\gM$, e.g. $(\sU^d_1,\dots,\sU^d_K)$}
    \IF {\text{Gromov}}
        \STATE $z_c = (x, w_c)_0 + b_c$\COMMENT{Calculate Gromov product per class as in \eqref{eq:gromov-product}}
    \ELSE
        \STATE $z_c = d_\gM(x, \tilde{H}_{a_c, p_c}^\gM)$\COMMENT{Calculate signed distance to the class hyperplane}
    \ENDIF
    \STATE $p(x) = \operatorname{softmax}(z_1, \dots, z_C)$ \COMMENT{Calculate Softmax for predictions}
    \end{algorithmic}
\end{algorithm}